\pdfoutput=1

\documentclass[11pt]{article}

\usepackage[]{acl}
\usepackage{times}
\usepackage{latexsym}

\usepackage{graphicx}
\usepackage{booktabs,caption}
\usepackage[flushleft]{threeparttable}
\usepackage{tabularx}
\usepackage{multicol}
\usepackage{multirow}

\usepackage[T1]{fontenc}

\usepackage[utf8]{inputenc}

\usepackage{microtype}

\title{Lexical Retrieval Hypothesis in Multimodal Context}

\author{Po-Ya Angela Wang, Pin-Er Chen, Hsin-Yu Chou, Yu-Hsiang Tseng, Shu-Kai Hsieh \\
         Graduate Institute of Linguistics, National Taiwan University \\ differe94nt@gmail.com, cckk2913@gmail.com, r10142008@ntu.edu.tw, \\  seantyh@gmail.com, shukaihsieh@ntu.edu.tw}

\begin{document}
\maketitle
\begin{abstract}

Multimodal corpora have become an essential language resource for language science and grounded natural language processing (NLP) systems due to the growing need to understand and interpret human communication across various channels. In this paper, we first present our efforts in building the first Multimodal Corpus for Languages in Taiwan (MultiMoco). Based on the corpus, we conduct a case study investigating the Lexical Retrieval Hypothesis (LRH), specifically examining whether the hand gestures co-occurring with \textit{speech constants} facilitate lexical retrieval or serve other discourse functions. With detailed annotations on eight parliamentary interpellations in Taiwan Mandarin, we explore the co-occurrence between \textit{speech constants} and non-verbal features (i.e., \textit{head movement}, \textit{face movement}, \textit{hand gesture}, and \textit{function of hand gesture}). Our findings suggest that while hand gestures do serve as facilitators for lexical retrieval in some cases, they also serve the purpose of information emphasis. This study highlights the potential of the MultiMoco Corpus to provide an important resource for in-depth analysis and further research in multimodal communication studies.
\end{abstract}



\section{Introduction}

Over the past decades, there has been a growing interest in multimodal corpus linguistic research \citep{paquot2021practical}, which focuses on the analysis and comprehension of information from diverse modalities, including speech, image, and gesture. To facilitate research in this field and other interdisciplinary studies, the creation of multimodal corpora, or collections of data from various modalities, has become more crucial. 

We thereby introduce the Multimodal Corpus for Languages in Taiwan (MultiMoco corpus), a newly released multimodal corpus that includes audio, video, gestural, and textual data involving various languages and discourse contexts. The MultiMoco corpus is comprised of recordings of realistic interactions taken in news and interpellation in parliament, where interviews and spontaneous speech take place. The synchronization of the audio files, video recordings, and gestures enables researchers to study the link between the various communication modes.  These data assist researchers in annotating information on the speakers, their actions, and the communication context. This corpus is designed for human communication and interaction-related research, such as conversation analysis, multimodal machine learning, and natural language processing.

To demonstrate the feasibility of the MultiMoco Corpus, we conduct a case study based on the parlimental interpellation clips in Taiwan Mandarin, aiming to validate the widely discussed Lexical Retrieval Hypothesis (hereafter, LRH) \citep{dittmann1969body,ekman1972hand,butterworth1978gesture,rauscher1996lrh}, which suggests that gesture and verbal disfluency tend to co-occur in spontaneous speech. 

More specifically, we take \textit{speech constants}, based on the framework of \citet{voghera2001speechconstant}, as indicators of potential verbal disfluency. We annotate one verbal feature (\textit{speech constants}) as well as four non-verbal features, including three forms of non-verbal expressions (\textit{head movement}, \textit{face movement}, \textit{hand gesture}) and \textit{functions of hand gesture}. With careful annotation, we attempt to answer research questions as follows: (1) Could we observe co-occurrences between \textit{speech constants} and gestures in the context of interpellation? (2) If there are co-occurrences with \textit{speech constants}, do hand gestures mainly play the role of priming lexical items? And (3) Do the hand gestures serve other functions regarding interlocutors and the entire discourse context?

To provide guidance on utilizing the MultiMoco Corpus to address multimodal research problems, we first review studies on the multimodal corpus, the multimodal annotation framework, and the LRH (Section \ref{sec:literature-review}).  Following this, we outline the data collection and annotation framework for the case study in Section \ref{sec:data-collection} and Section \ref{sec:data-annotation}. Next, we analyze if the non-verbal features co-occur with \textit{speech constants} (Section \ref{sec:result-overview}). The LRH mechanism is examined by identifying the co-occurrences between \textit{speech constants} and LRH-related/ non-LRH-related functions of hand gesture (Section \ref{sec:co-occur-funcs-of-hand}), along with the individual performances discussed in Section \ref{sec:co-occur-individual}. Section \ref{sec:conclusion} concludes the paper.

\section{Related Works}
\label{sec:literature-review}

\subsection{Multimodal corpus}

Communication, by nature, is multimodal \citep{carter2008linking}, and thereby constructing multimodal corpora affords researchers the opportunity to get a comprehensive understanding of the cognitive mechanisms underlying communication.
"Multimodal corpus" can be defined at varying degrees depending on its architecture \citep{allwood2008multimodal}. Generally speaking, it refers to an online repository of language and communication-related content that contains several modalities. In a narrower sense, it can be specified with audiovisual materials accompanied by annotations and transcriptions. 

Most earlier multimodal corpora are for specific purposes, to which \citet{paquot2021practical} refers as "specialized" corpora. For example, the Mission Survival Corpus \citep{mccowan2003modeling}, the Multimodal Meeting (MM4) Corpus \citep{mccowan2005towards}, and the VACE corpus \citep{chen2006vace} are all collected from meeting situations. Others are task-oriented corpora elicited in lab settings, such as the Fruit Carts corpus \citep{gallo2006software}, CUlture-adaptive BEhavior Generation for interactions with embodied conversational agents (CUBE-G) \citep{rehm2009creating}, and the spatial task-based dialogue corpus, SaGA \citep{lucking2010bielefeld}. Still, others include dyadic conversation in academic discourse (e.g., the Nottingham Multi-Modal Corpus (NMMC) \citep{knight2008nottingham} and the Pisa Audiovisual Corpus project \citep{camiciottoli2015pisa}), providing domain-specific multimedia materials for English for Specific Purposes (ESP) learners in higher education. 

Recent corpora attempt to be less specific and purpose-oriented, or they provide neurobehavioral perspectives to analyze the multimodal data. \citet{mlakar2017corpus} select 4 recordings of multiparty conversations in a talk show that are more spontaneous and cover various topics. In addition, Communicative Alignment of Brain and Behaviour (CABB) \citep{eijk2022cabb} builts a dataset on recordings of 71 pairs of participants discussing innovative, unconventional objects ("Fribbles") \citep{barry2014meet}, which provides pre-and-post behavioral and fMRI measurement information to investigate elicited conversational data from various dimensions.      

The majority of the mentioned databases are tailored for specific applications or constructed on smaller sets of data. Nonetheless, the MultiMoco corpus presented in this study is exceptional in incorporating video and audio recordings from ten public news channels and interpellation videos, which encompass a broader spectrum of languages and communication genres. This renders it an invaluable resource for investigating multilingual and multimodal communication, with the capacity to accommodate multidimensional annotations. The extensive size and diversity of the MultiMoco corpus create prospects for examining diverse aspects of human communication and pave the way for novel research directions.

\subsection{Multimodal annotation framework}

Various annotation frameworks have been proposed to encode labels for gesture forms and corresponding functions \citep{bavelas1992interactive,mcclave2000linguistic, kendon2004gesture, muller2004forms,allwood2005mumin, bressem201371}. According to \citet{debras2021prepare}'s proposal, "articulator" (e.g., hand or head), and "configuration of articulator" (e.g., head nod, wave, or turn) should be formally annotated. Functional annotation is to indicate co-verbal intentions of gestures. The Facial Action Coding System (FACS; \citealp{ekman1997facial,clark2020facial}), for facial expression annotations, and the Linguistic Annotation System for Gestures (LASG; \citealp{bressem201371}), for hand annotations, are both well-designed but complicated annotation systems. Annotation frameworks such as these can be time-consuming and challenging to achieve annotation agreement. \citet{debras2021prepare} suggests that coarse-grained annotations can benefit the onset of the research. 

We here review the annotation frameworks that will be adopted in the case study. Firstly, \textit{speech constants} will be annotated to examine the LRH evaluated by \citet{trotta2021gestures}, given that gestures tend to co-occur with verbal disfluency. Referring to the guidelines in \citet{voghera2001speechconstant}, four types of \textit{speech constants} (i.e., \textit{pause}, \textit{repetition}, \textit{truncation}, and \textit{semi-lexical}) are taken as the annotation targets. Secondly, the non-verbal target features comprise forms and functions, namely \textit{head movement}, \textit{face movement} (\textit{eyebrows and mouth}), \textit{hand gesture}, and \textit{functions of hand gesture}. Considering \citet{debras2021prepare}'s suggestions for coarse-grained annotations, this study follows the concise annotation framework adopted by \citet{camiciottoli2015pisa}, incorporating gesture form abbreviations by \citet{julian2011evaluation} and the gesture functions by \citet{kendon2004gesture} and \citet{weinberg2013instructor}. In \citet{camiciottoli2015pisa}'s framework, \textit{head movement} include \textit{head-nodding/tilting/jerking/moving} together with multiple directions and repetition; \textit{face movement} involve the movement of eyebrows and mouth; \textit{hand gesture} mark the movements of fingers, palm, and the whole hand. The comprehensive labels and definitions for each feature will be explained in Section \ref{sec:data-annotation}.

\subsection{Lexical Retrieval Hypothesis}
As reviewed in \citet{ozer2020gesture}, multimodal interaction in speech production and comprehension regarding individuals' cognitive tendencies has been heatedly discussed. When a speaker cannot clarify intended thoughts, gestures are incorporated during hesitation pauses or the lexical pre-planning stage \citep{dittmann1969body, butterworth1978gesture}. The link between verbal, non-verbal, and conceptual aspects can be addressed by the "growth point," the smallest thought unit, comprising both utterances and gestures \citep{mcneill1992hand}. \citet{krauss1998we} has considered the relationship between thoughts, utterances, and gestures from another perspective, specifying three parts in speech production: conceptualizing, grammatical encoding, and phonological encoding. Among these three parts, phonological encoding, the retrieval of lexical form, is the part where gestures affect     the verbal modality, and limited gestures reduce speech fluency when a speaker discusses spatial information \citep{krauss1998we}. Later, \citet{krauss1999role} have further proposed that concepts in the mind are stored in various forms, so activating one idea in one modality may also activate concepts in other modalities. Thus, concepts can be fully comprehended when information from different modalities is all presented, and representations from one modality can be converted into another modality. Following the line of this discussion, the gestural modality can assist lexical retrieval in the verbal modality because of such cross-modal priming. This is termed the "Lexical Retrieval Hypothesis" \citep{gillespie2014verbal,trotta2021gestures}. Namely, LRH refers to the process that the triggered idea's lexical gestures\footnote{\citet{krauss1998we} refers to these lexical retrieval supporting gestures as "lexical gestures."} (i.e., gestures that can iconically represent meanings) can semantically prime the phonological encoding of the related words \citep{gillespie2014verbal}. \citet{gillespie2014verbal} also specify that LRH is less applicable if the speaker can resort to alternative tactics to avoid lexical access challenges, which occur in improvisational speech production.

The Lexical Retrieval Hypothesis is tested in several tasks and contexts. \citet{hostetter2007raise} distinguish the phonemic fluency from the semantic fluency\footnote{"Phonemic fluency" indicates thought-organizing skills associated with representational gesture rates, whereas "semantic fluency" is less correlated with representational gesture rates but has a significant correlation with beat gestures,}, suggesting lexical access efficiency may be related to different types of gestures. Additionally, \citet{smithson2013verbal} have proposed that the negative association between verbal working memory and iconic gesture production in bilinguals designates gesture production's assistance in the retention and utilization of language information. \citet{trotta2021gestures} calculate the weighted mutual information (WMI) between the hand movements and the concurrent speech disfluency features involving five kinds of \textit{speech constants}\footnote{Five kinds of \textit{speech constants}: pause, repetition, truncation, and semi-lexical), as specified by \citet{voghera2001speechconstant}}. The result concurs with the LRH since hand gestures are more related to semi-lexical features and pauses in interview contexts. It is noted that in \citet{trotta2021gestures}, \textit{speech constants} are considered disfluency features to assess the LRH, whereas hesitation pauses may signal lexical retrieval difficulties. 

As most of the studies mentioned have examined the LRH with laboratory tasks or free-form interviews, we aim to assess the LRH in formal speaking contexts (i.e., political interpellation) as well as its applicability in less colloquial speech. This case study conjectures that gestures co-occurring with \textit{speech constants} are not just for facilitating lexical retrieval.

\section{Methodology}
Our study of the lexical retrieval hypothesis is based on the multimodal data made available from Multimoco. We first introduce the construction and contents of the MultiMoco Corpus (Section \ref{sec:moco}). Then, the data collection for our case study on the LRH is illustrated (Section \ref{sec:data-collection}), followed by the annotation framework for the target features (Section \ref{sec:data-annotation}). The annotation results and analyses will be discussed in the subsequent sections.

\subsection{MutliMoco Corpus}
\label{sec:moco}
The MultiMoco Corpus is built on recorded videos and audios from 10 public television channels\footnote{The target channels are as follows: CTV News PTS News, PTS Taigi, Hakka TV, Taiwan Indigenous TV, TTV News, CTS News, Congress Channel I, Congress Channel II, and FTV News.} in Taiwan, including news in multiple languages (i.e., Taiwan Mandarin, Taiwan Southern Min, Hokkien, Hakka, and Formosan languages) and the interpellation of the Taiwan Legislative Yuan (the parliament of Taiwan). While the TV news is recorded by wireless television receivers, the interpellation video clips with transcriptions in Taiwan Mandarin are retrieved directly from the Internet Multimedia Video-on-Demand System for Rebroadcasting Legislative Yuan Proceedings\footnote{https://ivod.ly.gov.tw/Demand}. 

\begin{figure}[!ht]  
    \includegraphics[width=\columnwidth]{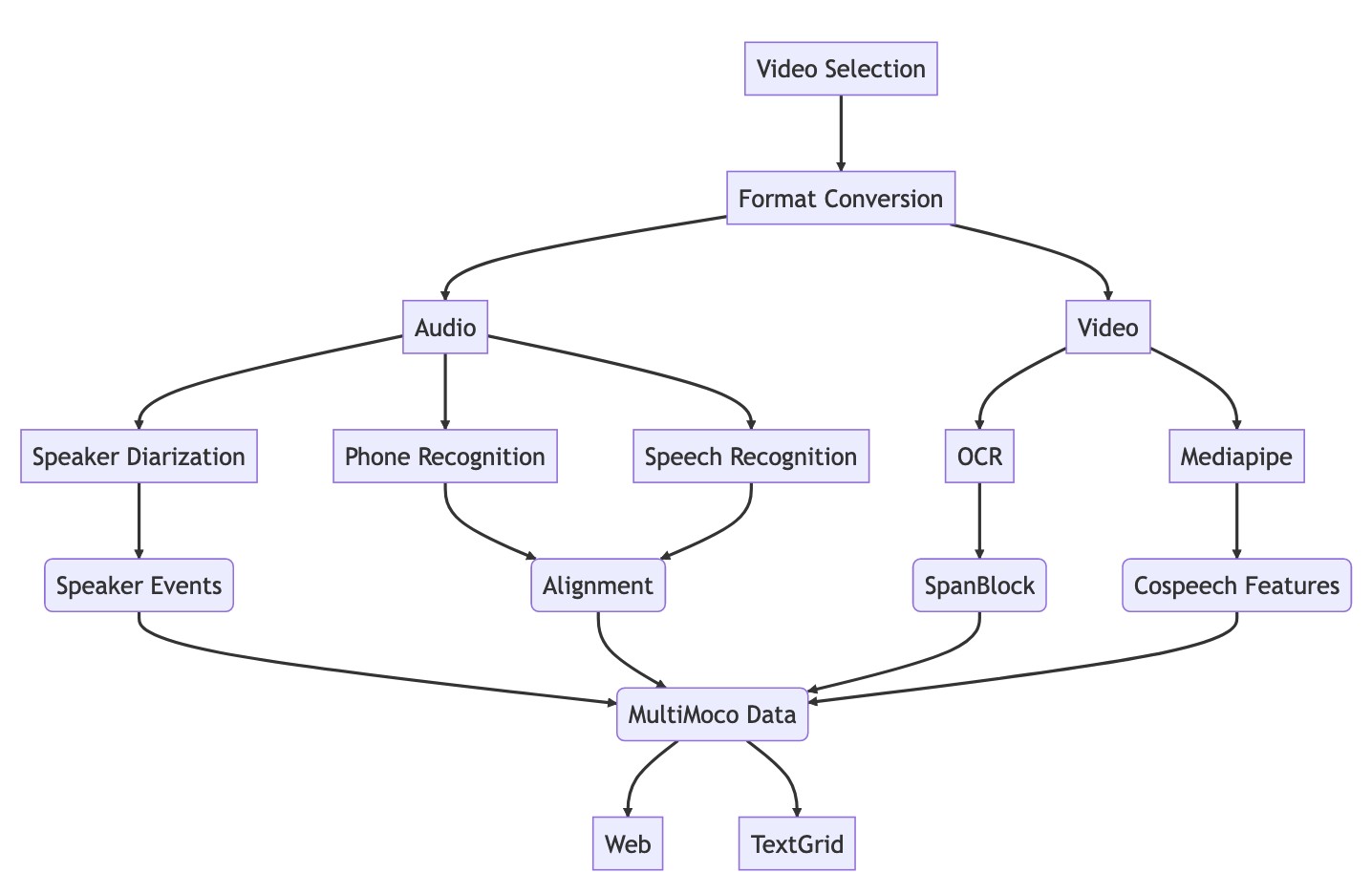}
    \caption{Establishment workflow of the MultiMoco Corpus}
    \label{fig:moco-workflow}
\end{figure}

Figure \ref{fig:moco-workflow} displays the data processing workflow of the MultiMoco Corpus. With 223 video clips from Taiwan public television channels and the interpellation from Taiwan Legislative Yuan, the MultiMoco Corpus provides 5,854 minutes of dialogue, accompanied by 1,485,297 characters of captions transcribed via Whisper \citep{radford2022whisper} model. In addition, 22,805 gestures identified via MediaPipe \citep{lugaresi2019mediapipe} are also included in the corpus. The multimodal nature of the corpus allows researchers to conduct cross-modality analyses, thereby broadening the understanding of the communicative potential of various modalities beyond spoken texts. That is, the MultiMoco Corpus provides us with the potential to extend communication studies to diverse linguistic and multimodal contexts.

\subsection{Data collection}
\label{sec:data-collection} 
\begin{figure}[!ht]  
    \includegraphics[width=\columnwidth]{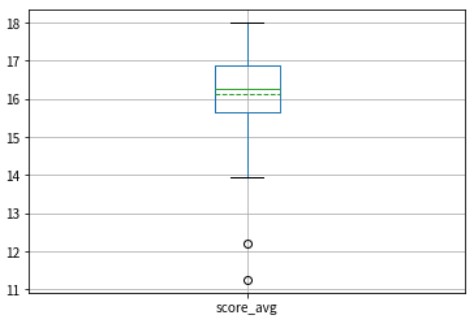}
    \caption{Descriptive statistics of citizen evaluation score}
    \label{fig:citizen_evaluation}
\end{figure}

Our lexical retrieval analysis data are extracted from MultiMoco Corpus, specifically focusing on spontaneous speech during interpellation involving interactions between legislators and officers. We have selected interpellation clips based on the evaluation scores of 103 legislators from the Citizen Congress Watch (CCW) in the 10th session of Congress. Figure \ref{fig:citizen_evaluation} shows the distribution of individually-averaged evaluation scores, with an average score of approximately 16, a minimum of 11.25, and a maximum of 17.998. After considering the evaluation score, interpellation topics, and political parties, we choose four legislators (two with higher evaluation scores and two with lower evaluation scores) for subsequent multimodal analyses. We collect eight interpellation clips, each lasting between 8 and 12 minutes and featuring a male and a female legislator in each pair. The interpellation topics are detailed in Table \ref{tab.clip_topics}.

\begin{table}[!h]
\centering
\begin{tabular}{p{1mm}p{15mm}p{24mm}} 
\toprule 
\multicolumn{1}{c}{\textbf{Legislator}} & \multicolumn{2}{c}{ \textbf{Topic of Interpellation Clips}}  \\
\midrule
high\_A  & Social welfare &  Education and culture \\
high\_B  & Finance        & Communications  \\
low\_C   & Finance        & Judiciary and organic laws \\
low\_D   & Social welfare &  Education and culture  \\
\bottomrule 
\end{tabular}
\caption{Topics of the interpellation clips. The prefixes (\textit{high} or \textit{low}) in the Legislator column are used for identifying the evaluation scores for the legislators (i.e., A, B, C, and D).}
\label{tab.clip_topics}
\end{table}

\begin{table}[!ht]
\centering
\small
\begin{tabularx}{\columnwidth}{lX}
\toprule 
\multicolumn{1}{c}{\textbf{Label}} & \multicolumn{1}{c}{\textbf{Definition}}  \\
\midrule
\texttt{Pause} & This marks a pause either between or within utterances. \\
\texttt{Repetition} & This marks interjections (e.g., eh and ehm), or more general words that convey the meaning of an entire sentence, constituting a complete linguistic act demonstrated by their paraphrasability. \\
\texttt{Non-lexical item} & This marks cases of repetition of utterances in order to give coherence and cohesion to the speech or self-repetition as a control mechanism of the speech programming. \\
\texttt{Truncation} & This indicates the deletion of a phoneme or a syllable in the final part of a word. \\
 \bottomrule 
\end{tabularx}
\caption{Labels for speech constants. It is noted that the original label ``semi-lexical'' in \citet{trotta2021gestures} is renamed ``non-lexical item'' in our study.}
\label{tab.speech_constant}
\end{table}

\subsection{Data annotation framework}
\label{sec:data-annotation}
We investigate the functions of non-verbal features and their co-occurrence with disfluency in spontaneous speech. Three non-verbal forms (i.e., \textit{head movement}, \textit{face movement}, and \textit{hand gesture}), one non-verbal function (i.e., \textit{functions of hand gesture} ), and one verbal feature (i.e., \textit{speech constants}) are selected as our annotation targets; the latter is used to identify disfluency in speech. 

Considering the specificity of each feature and the consensus in prior studies, we adopt different annotation frameworks for corresponding features. The \textit{speech constants} are annotated based on the framework in \citet{voghera2001speechconstant}, as shown in Table \ref{tab.speech_constant}; \textit{functions of hand gesture} were annotated via Camiciottoli and Bonsignori’s framework, as presented in Table \ref{tab.functions-hand}. The three non-verbal forms (i.e., \textit{head movements}, \textit{face movements}, and \textit{hand gestures}) are classified based on \citet{camiciottoli2015pisa}'s framework, as illustrated in Table \ref{tab.cospeech_ges}. It is noted that the labels in the table are generalized to a more coarse-grained scale regarding the entailment of the original labels. 
 
Five native speakers annotate the five verbal and non-verbal features (i.e., \textit{head movement}, \textit{face movement}, \textit{hand gesture}, \textit{function of hand gesture}\footnote{For clarity, we use the \textit{italic} form when referring to the five targets, and we use the \texttt{typewriter} font when referring to the labels under each target.}, and \textit{speech constants}) via ELAN \citep{sloetjes2008elan}\footnotemark, an open-source software appropriate for multimodal annotations and linguistic analysis. Take \textit{speech constants} for instance, the two annotators separately mark the time periods and corresponding labels of \textit{speech constants} that occur in all eight clips. Then, the annotated pair of tiers (made by the two annotators) for each clip are segmented into units of 100 milliseconds and aligned with each other. 

For annotation consistency, the annotators are asked to annotate different features from clip segments and decide on an agreed-upon criterion for disagreed annotations. For instance, the function, \texttt{Parsing}, marks situations in which a speaker intends to initiate a new discourse turn, recur the same gesture as if beating, or make some trivial movements that have no clear reference. As we focus on the co-occurrence and association between non-verbal features and disfluency, we will not inspect the details of the annotation results within each non-verbal feature but rather discuss the general co-occurrence with \textit{speech constants} in the following sections.

\footnotetext{ELAN (https://archive.mpi.nl/tla/elan); Max Planck Institute for Psycholinguistics, The Language Archive, Nijmegen, The Netherlands.}

\begin{table}[!h]
\small
\begin{tabularx}{\columnwidth}{ll}
\toprule 
\multicolumn{1}{c}{\textbf{Label}} & \multicolumn{1}{c}{\textbf{Definition}}  \\
\midrule
\texttt{Social} & social (emphasizing a message) \\
\texttt{Repres} & representational (representing object/idea) \\
\texttt{Index} &  indexical (indicating a referent) \\
\texttt{Parsing} &  parsing (distinguishing units of speech) \\ 
\texttt{Perform} &  performative (illustrating speech act) \\
\texttt{Modal} &  modal (expressing certainty/uncertainty) \\
 \bottomrule 
\end{tabularx}
\caption{Labels for functions for hand gesture. The functions of `beat' and `representational' gestures in \citet{hostetter2007raise} are represented as \texttt{Parsing} and \texttt{Representational} in this study.}
\label{tab.functions-hand}
\end{table}

\begin{table}[!h]
\centering
\small
\begin{tabularx}{\columnwidth}{ll}
\toprule 
\multicolumn{1}{c}{\textbf{Type}} & \multicolumn{1}{c}{\textbf{Label}}  \\
\midrule
Face & 
    \begin{tabular}[t]{@{}l@{}}Finger pointing towards audience\\ Hands sweeping sideways\\ Hands rotating at center of body\\ Hands wide apart moving down\\ Hands clasped together in front of body\\ Other\end{tabular} \\
\midrule
Head & 
    \begin{tabular}[t]{@{}l@{}}Nod\\ Jerk\\ Move Forward/Backward\\ Tilt\\ Side-turn\\ Shake (repeated)\\ Other\end{tabular} \\
\midrule
Hand & 
    \begin{tabular}[t]{@{}X@{}}Finger pointing towards audience\\ Hands sweeping sideways\\ Hands rotating at center of body\\ Hands wide apart moving down\\ Hands clasped together in front of body\\ Other\end{tabular} \\
 \bottomrule 
\end{tabularx}

\caption{Labels for co-speech gestures: face, head, and hand}
\label{tab.cospeech_ges}
\end{table}

\begin{figure*}[!ht]
    \includegraphics[width=0.8\textwidth]{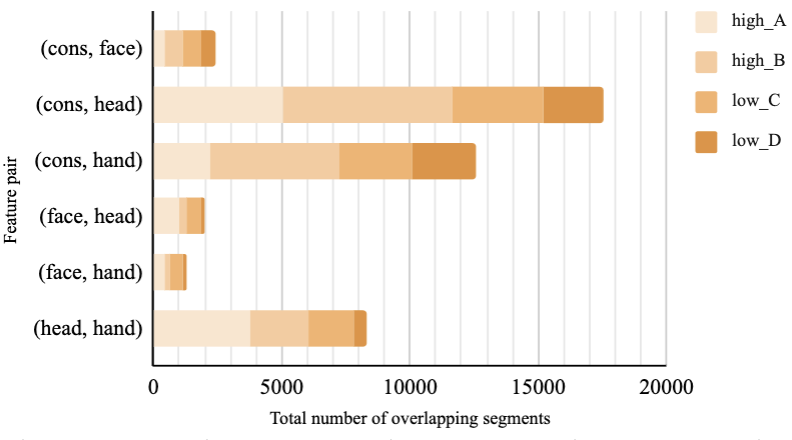}
    \caption{Co-occurrences of different feature pairs. The y-axis represent the number of overlapping segments between different pairs of annotated features.}
    \label{fig:cooccur-overview}
\end{figure*}

\section{Results \& Discussions}

We first examine the non-verbal features' co-occurrence with \textit{speech constants}, which indicate verbal disfluency (Section \ref{sec:result-overview}). Then, the potential discourse \textit{functions of hand gesture} will be analyzed (Section \ref{sec:co-occur-funcs-of-hand}). Finally, we will discuss more comprehensive gesture functions independent of verbal disfluency but related to interlocutors and the entire discourse context in Section \ref{sec:co-occur-individual}.

\subsection{Co-occurrence overview}
\label{sec:result-overview}
As we target one verbal feature (\textit{speech constants}) and three forms of non-verbal features (head, hand, and face)\footnote{It should be noted that one non-verbal related feature, i.e., the \textit{functions of hand gesture}, are annotated based on the occurrence of hand gesture; thus, calculating the co-occurrences (i.e., overlapping segments) between \textit{functions of hand gesture} and the other features would be meaningless, as it would be the same as hand gesture.}, we calculate the co-occurrences\footnote{The co-occurrence of one pair of features is defined as the summed number of overlapping segments; one segment is a unit of 100 milliseconds.} of the six patterns by modality. Figure \ref{fig:cooccur-overview} shows that \textit{head movement} and \textit{speech constants} co-occur most frequently, followed by \textit{hand} and \textit{speech constants}. \textit{Face movement} shows fewer co-occurrences with the other features (i.e., face \& head, face \& hand, and face \& \textit{speech constants}), which may relate to the few occurrences of face movement in all clips. While the non-verbal features tend to co-occur with one another, the frequencies are far lower than their respective co-occurrence with \textit{speech constants}. This may correspond to the LRH that when \textit{speech constants} appear, i.e., during hesitation pauses or the lexical pre-planning stage, non-verbal gestures are possibly employed by the speaker as well \citep{dittmann1969body, butterworth1978gesture}.

\subsection{ Co-occurring functions of hand gestures }
\label{sec:co-occur-funcs-of-hand}

\begin{table*}[!ht]
\begin{tabularx}{\textwidth}{p{28mm}p{16mm}p{13mm}p{22mm}p{29mm}p{11mm} | p{6mm}}
\toprule
(SC / FH) & \textbf{Indexical} & \textbf{Parsing} & \textbf{Performative} & \textbf{Representational} & \textbf{Social} & \textbf{Total} \\
\midrule
\textbf{Non-lexical item} & 10 & 24 & 2  & 16 & 9   & 61           \\
\textbf{Pause}            & 59 & 87 & 20 & 46 & 133 & \textbf{345} \\
\textbf{Repetition}       & 6  & 16 & 0  & 7  & 32  & 61           \\
\textbf{Truncation}       & 8  & 0  & 0  & 0  & 3   & 11          \\
\midrule
\textbf{Total}            & 83 & 127 & 22 & 69 & 177 & 478        \\
\bottomrule
\end{tabularx}
\caption{Contingency table of \textit{speech constants} and \textit{functions of hand gesture}. SC represents \textit{speech constants}, and FH represents \textit{functions of hand gesture}.
}
\label{tab:speechconstants-function}
\end{table*}

\begin{figure*}[!h]  
    \includegraphics[width=\textwidth]{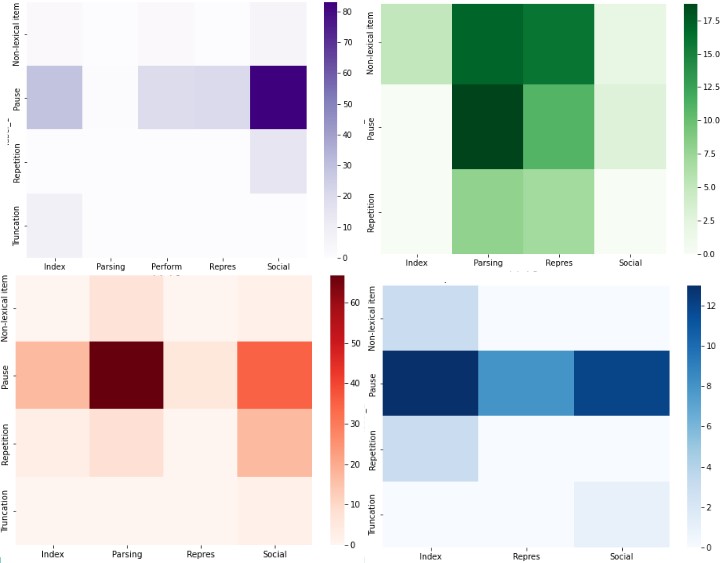}
    \caption{Heat maps of co-occurrence of \textit{speech constants} and \textit{functions of hand gesture} (by-legislator). The upper left image belongs to high{\_C}, the upper right image belongs to high{\_B}, the lower left image belongs to high{\_D}, and the lower right image belongs to high{\_A}.}
    \label{fig:cons-func}
\end{figure*}

 As significant as the respective gesture co-occurrence with \textit{speech constants} is, could we claim that the identified \textit{speech constants} require gestures to facilitate lexical retrieval? To further understand the purposes of the hand gestures co-occurring with \textit{speech constants}, Table \ref{tab:speechconstants-function} below presents the overall frequencies of each type of \textit{speech constants} co-occurring with different \textit{functions of hand gesture}. \textit{Speech constants}, especially \texttt{non-lexical items} and \texttt{pauses}, are taken as verbal disfluency traits in the LRH evaluation \citep{trotta2021gestures}. We would like to argue that the intentions of performing \textit{speech constants} are various, so the functions resulting from the interplay between verbal and non-verbal modalities are complicated. Thus, in addition to using \textit{speech constants} as markers of the possible presence of verbal disfluency, we study the functions of co-occurring hand gestures in order to realize whether the co-occurring hand gestures are lexical retrieval facilitators or carry out other functions in speech contexts. 
 
First, we examine the distributions of \textit{speech constants} and their co-occurring \textit{functions of hand gesture}. Regarding \textit{speech constants}, \texttt{pause} is the most frequently observed category with 345 frequencies, accounting for 72.2{\%} co-occurrences among all. \texttt{Repetition} and \texttt{non-lexical item} both rank second. \texttt{Truncation} sporadically occurs in the collected dataset. As for \textit{functions of hand gesture}, \texttt{Social} (i.e., to emphasize a message) is the most frequent function for the \textit{speech constants} as a whole. The rest of the ranking goes as follows: \texttt{Parsing} > \texttt{Indexical} > \texttt{Representational} > \texttt{Performative} > \texttt{Modal}\footnote{As there is no co-occurrence between \texttt{Modal} and \textit{speech constants}, this label is not display in Table \ref{tab:speechconstants-function}.}.

\citet{trotta2021gestures} claim that more hand gestures go with semi-lexical items (``non-lexical item'' in our study) and that \texttt{pauses} can confirm LRH. In this way, if we take \textit{speech constants} as the speech disfluency indicators, then \texttt{pauses} and \texttt{non-lexical items} seem to be the focused indicator to evaluate the LRH. In the following analysis, we focus on \textit{function of hand gesture} co-occurring with \texttt{pause} and \texttt{non-lexical item}. These functions of concurrent hand gesture can be subcategorized into LRH-related functions (\texttt{Parsing} and \texttt{Representational}) and non-LRH-related functions (\texttt{Social}, \texttt{Indexical}, and \texttt{Performative}), for beat and representational gestures receptively correlate with different types of fluency \citep{hostetter2007raise}.

Starting from the LRH-related functions of hand gesture, Table \ref{tab:speechconstants-function} shows that \textit{functions of hand gesture} co-occurring with \texttt{pause} and \texttt{non-lexical item} account for 42.6{\%}. \texttt{Parsing} is the second-highest intended \textit{function of hand gesture} co-occurring with \texttt{pause}; this is noticeably consistent with the obvious correlation between semantic fluency and beat gestures \citep{hostetter2007raise}. Although \texttt{pauses} co-occur with hand gestures for \texttt{Representational} rank fourth, it still comprises 13.3{\%} of total occurrences. In the case of \texttt{non-lexical items}, hand gestures for \texttt{Parsing} and \texttt{Representational} functions show higher frequencies for appearing with \texttt{non-lexical item} (65.5{\%}), suggesting that hand gestures co-occurring with \texttt{non-lexical item} are more likely to facilitate verbal delivery in formal speech. From the discussion above, it can be concluded that \texttt{pauses} and \texttt{non-lexical items} are often accompanied by hand gestures for \texttt{Parsing} and \texttt{Representational}, which primes lexical retrieval.

When it comes to non-LRH-related functions of concurrent hand gestures, pause is highly associated with hand gestures for \texttt{Social} function. This indicates that \texttt{pauses} seem not primarily to represent hesitation pauses, but rather to emphasize the primary topic of the speech in interpellation. Subsequently, \texttt{Indexical} is the ranked third \textit{function of hand gestures} synchronizing with \texttt{pause}, implying that speakers prefer to depict the referent with visual-motion modality. \texttt{Performative} function is the least frequent one, but its occurrence is still significant compared to other \textit{speech constants}. \texttt{Indexical} function in \texttt{non-lexical item} case is subtly higher than \texttt{Social} and \texttt{Performative}. As shown in Figure\ref{fig:cons-func}, it can be inferred that synchronous hand gestures of \texttt{pause} and \texttt{non-lexical item} also carry out information emphasis and referent depiction functions.

To sum up, in formal speech hand gestures co-occurring with \textit{speech constants} related to speech disfluency are not just used to iconically represent the unspoken thoughts but also serve the function of reinforcing the verbal information.  

\subsection{Co-occurrence of individual legislators}
\label{sec:co-occur-individual}

This research takes formal speech as research target to reexamine the applicability of LRH in individual performance since \citet{gillespie2014verbal} specify that LRH is less applicable if the speaker can use alternate strategies to circumvent lexical access difficulties that arise during improvised speech. \citet{trotta2021gestures} illustrate that LRH does not confirm in all interviewers' performances, whereas the applicability of LRH in formal speech stays unclear. Accordingly, the purpose of this section is to highlight the functions adopted by all speakers and their implications related to LRH.

According to individual speaker behaviors in Figure \ref{fig:cons-func}, \texttt{Social}, \texttt{Indexical}, and \texttt{Representational} are the functions employed by all of speakers. This exemplifies that information accentuation and referent portrayal are primary functions of the synchronous hand gestures despite possible variations in individual style preferences. Notably, all speaker adopt the concurrent hand gestures for the \texttt{Representational} function when pausing, indicating the widespread use of nonverbal modalities to compensate for verbal delivery difficulties in improvised speech situations. This offers a new perspective to extend the suggestions presented by \citet{gillespie2014verbal}, highlighting the general applicability of hand gestures to serve the lexical retrieval purpose in formal spontaneous speech contexts.

\section{Conclusion}
\label{sec:conclusion}
 
In conclusion, this paper highlights the creation of a multimodal corpus of Taiwanese languages and its application in investigating the lexical retrieval hypothesis in hand gestures and speech. 

The case study presented in this paper examines the application of multimodal corpora in the investigation of the lexical retrieval hypothesis, indicating that hand gestures often accompany \textit{speech constants} such as \texttt{pauses} and \texttt{non-lexical items}, priming the function of lexical retrieval. By leveraging the corpus, our finding suggests that hand gestures are not solely for retrieval struggles but can also serve as means of emphasizing information. Additionally, the outcome of individual speech performances signifies the general applicability of hand gestures for the lexical retrieval purpose. We believe that the continued development and utilization of this corpora will pave the way for enhancing our understanding of the intricate interplay between verbal and non-verbal communication channels.

\label{sec:bibtex}


\bibliography{anthology,custom}

\end{document}